\documentclass[runningheads]{llncs}
\usepackage{graphicx}

\usepackage{tikz}
\usepackage{comment}
\usepackage{amsmath,amssymb} 
\usepackage{orcidlink}
\usepackage{color}

\usepackage[accsupp]{axessibility}  

\usepackage{comment}
\usepackage[switch]{lineno}
\usepackage{url}            
\usepackage{booktabs}       
\usepackage{amsfonts}       
\usepackage{nicefrac}       
\usepackage{microtype}      
\usepackage{xcolor}         
\usepackage{subcaption}
\usepackage{csquotes}
\usepackage{amsmath}
\usepackage{multirow}
\usepackage{verbatim} 
\usepackage{diagbox}
\usepackage{bm}
\usepackage{makecell}
\usepackage{newfloat}
\usepackage{caption}
\usepackage{hyperref}
\newcommand{\ie}{i.e., }
\newcommand{\eg}{e.g., }

\begin{document}
\pagestyle{headings}
\mainmatter
\def\ECCVSubNumber{2686}  

\title{Selective Query-guided Debiasing for Video Corpus Moment Retrieval} 

\titlerunning{Selective Query-guided Debiasing Network}
%
\author{Sunjae Yoon\inst{1}\orcidlink{0000-0001-7458-5273} \and
Ji Woo Hong\inst{1}\orcidlink{0000-0002-3758-0307} \and
Eunseop Yoon\inst{1}\orcidlink{0000-0002-5580-5354} \and
Dahyun Kim\orcidlink{0000-0003-0881-8651} \and
Junyeong Kim\inst{2}\orcidlink{0000-0002-7871-9627} \and
Hee Suk Yoon\inst{1}\orcidlink{0000-0003-2115-8459} \and
Chang D. Yoo\inst{1}\thanks{Corresponding Author}\orcidlink{0000-0002-0756-7179}}
\authorrunning{S. Yoon et al.}
%
\institute{Korea Advanced Institute of Science and Technology, Daejeon, Republic of Korea \and
Chung-Ang University, Seoul 06974, Republic of Korea\\
\email{sunjae.yoon@kaist.ac.rk, junyeongkim@cau.ac.kr, cd\_yoo@kaist.ac.kr}}
\maketitle

\begin{abstract}
Video moment retrieval (VMR) aims to localize target moments in untrimmed videos pertinent to a given textual query. Existing retrieval systems tend to rely on retrieval bias as a shortcut and thus, fail to sufficiently learn multi-modal interactions between query and video. This retrieval bias stems from learning frequent co-occurrence patterns between query and moments, which spuriously correlate objects (e.g., a pencil) referred in the query with moments (e.g., scene of writing with a pencil) where the objects frequently appear in the video, such that they converge into biased moment predictions. Although recent debiasing methods have focused on removing this retrieval bias, we argue that these biased predictions sometimes should be preserved because there are many queries where biased predictions are rather helpful. To conjugate this retrieval bias, we propose a Selective Query-guided Debiasing network (SQuiDNet), which incorporates the following two main properties: (1) Biased Moment Retrieval that intentionally uncovers the biased moments inherent in objects of the query and (2) Selective Query-guided Debiasing that performs selective debiasing guided by the meaning of the query. Our experimental results on three moment retrieval benchmarks (i.e., TVR, ActivityNet, DiDeMo) show the effectiveness of SQuiDNet and qualitative analysis shows improved interpretability. The project is available at \href{https://github.com/dbstjswo505/SQuiDNet}{\texttt{github.com/dbstjswo505/SQuiDNet}}.

\keywords{video moment retrieval, retrieval bias, selective debiasing}
\end{abstract}

\section{Introduction}

Video streaming services (e.g., YouTube, Netflix) have rapidly grown these days, which promotes the development of video search technologies.
As one of these video search technologies, video moment retrieval (VMR) \cite{gao2017tall,anne2017localizing} serves as essential building block to underpin many frontier interactive AI systems, including video/image captioning \cite{yu2016video,Venugopalan_2015_ICCV}, video/image question answering \cite{tapaswi2016movieqa,lei2018tvqa} and visual dialog \cite{das2017visual}.
VMR aims to localize temporal moments of video pertinent to textual query.
Recently, the growing interest in video searching drove this VMR to perform in a more general format of retrieval, referred to as video corpus moment retrieval (VCMR) \cite{lei2020tvr}. 
VCMR also aims to localize a moment like VMR, but the search spaces extend to a `video corpus' composed of the large number of videos. 
Therefore, given query, VCMR conducts two sub-tasks: (1) identifying relevant video in the video corpus and (2) localizing a moment in the identified video. 
Despite this respectful effort to generalize video retrieval, the VCMR systems still suffer from dependence on retrieval bias, which hinders the system from accurately learning multi-modal interactions.
%
\begin{figure}[t]
    \centering
    \includegraphics[width=0.76\linewidth]{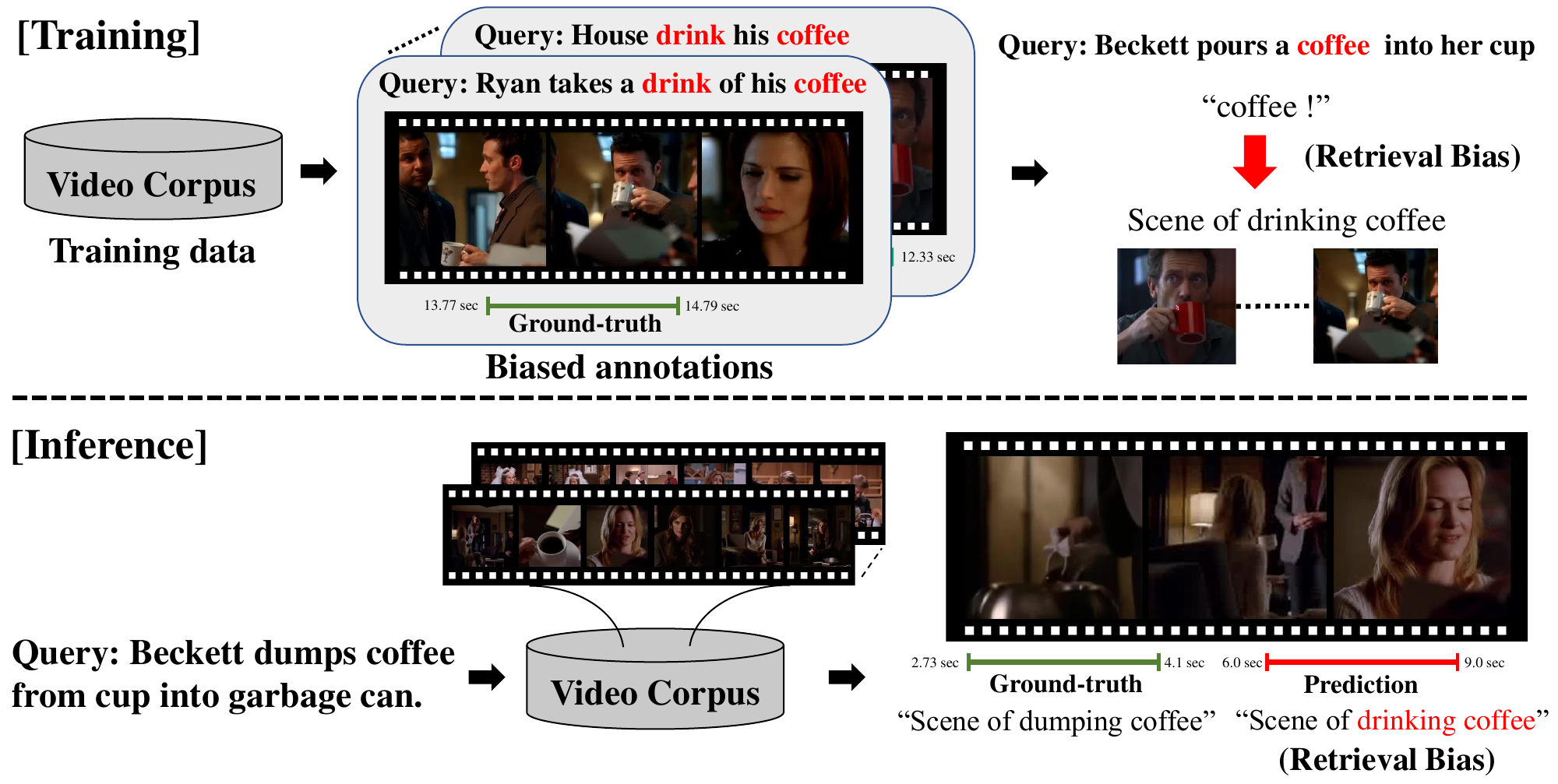}
    \caption{VCMR training and inference. The biased annotations in training dataset make retrieval bias, which causes biased moment prediction in the inference.}
    \label{fig:1}
\end{figure}
Figure \ref{fig:1} gives an example of incorrect moment predictions due to the retrieval bias.
Given query as ``Beckett dumps coffee from cup into garbage can" in inference time, current retrieval systems make incorrect moment prediction with the scene of `drinking a coffee'.
This is because annotations of training dataset include many co-occurrences between the object word `coffee' in query and the scene of `drinking,' which leads to biased moment prediction referred to as \textit{retrieval bias}. 
This retrieval bias constrains an object (e.g., coffee) to specific scene (e.g., scene of drinking), thus the other scenes related to that object word lose chance to be searched.
Recent debiasing methods \cite{nan2021interventional,yang2021deconfounded} have focused on removing or mitigating this retrieval bias as they assume the bias degrades retrievals.
However, we argue that these biased predictions sometimes should be preserved because there are many queries where biased prediction is rather helpful, such that selective debiasing is required.
\begin{figure}[t]
    \centering
    \includegraphics[width=\textwidth]{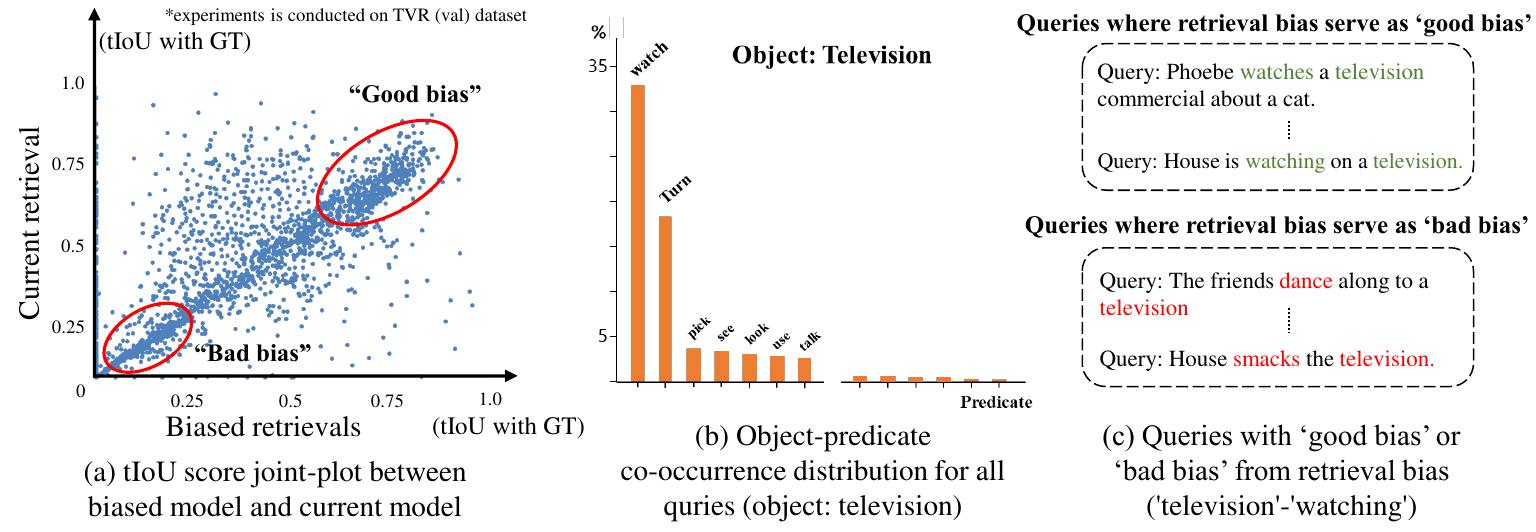}
    \caption{(a) All predictions' tIoU score joint plot between biased model and current model shows correlations between two models, (b) object (`television')-predicate co-occurrence distribution for all queries shows predominant predicate word (`watch'), (c) exemplifies queries where the retrieval bias (`television'-`scene of watching television') serves as `good bias' or `bad bias' from statistics in (b).}
    \label{fig:2}
    \vskip -0.2in
\end{figure}

Our experimental studies in Figure \ref{fig:2} prove that retrieval bias can also be `good'.
Figure 2-(a) presents a temporal intersection of union (tIoU) scores for all queries between moment predictions and ground-truth.
The predictions are from two retrieval models: (1) the current best performance model and (2) the biased retrieval model.
The biased retrieval model is intended to predict biased moments for given queries. 
To implement this, we simply build a toy model and give it `nouns in the query' as inputs instead of `full query sentence'. 
This induces a deficiency of the original query's meaning and leads the model to depend on predicting moments where those nouns are mainly used.
After that, we represent these two retrieval models' predictions in a joint-plot of tIoU scores, where it is noted that the plot shows positive correlations.
This correlation stands out strong in predictions of low and high tIoU scores, which tells that the current retrieval is both harmed and helped by the retrieval bias.
%
%
Therefore, retrieval bias includes both \textit{good bias} and \textit{bad bias}.
Then, what distinguishes good bias and bad bias?
Figure \ref{fig:2}(b) give example of our insight on this.
We investigate predicates that appear together with a specific word (\eg television) in all queries and identify that one or two predicates (\eg watch, turn) are predominantly bound with that word.
From these, we have knowledge that query sentences including object word and its most co-occurrent predicate should benefit from retrieval bias (i.e., good bias) because there are also many corresponding scenes (e.g., scene of watching television), but queries with other predicates would be degraded (i.e., bad bias) by this retrieval bias.
Figure \ref{fig:2}(c) shows query samples where the retrieval bias (\ie television-scene of watching) serves as `good bias' or `bad bias'.

Intrigued by these two characteristics of retrieval bias, we propose a Selective Query-guided Debiasing network (SQuiDNet), which incorporates the following two main properties: (1) Biased Moment Retrieval (BMR) that intentionally uncovers the retrieval bias inherent in objects of the query and (2) Selective Query-guided Debiasing (SQuiD) that performs selective debiasing via disentangling `good' and `bad' retrieval bias according to the meaning of the query.
In the overall pipeline, we first prepare two moment retrieval models: (1) Naive Moment Retrieval (NMR) and (2) Biased Moment Retrieval (BMR), where both predict the start-end-time of the moment pertinent to their input queries.
The NMR is trained under the original purpose of VCMR, so it takes a video and query pair as inputs and sufficiently learns video-language alignment.
However, the BMR is trained under the motivation of learning retrieval bias, so it takes a video and `object words' in the query instead of a full query sentence.
These words lose the contextual meaning of the original query, which makes the BMR difficult to properly learn a vision-language alignment, and rather, depend on the shortcut of memorizing spurious correlations that link given words to specific scenes.
Based on these two retrievals, SQuiD decides whether the biased prediction of BMR is `good bias' or `bad bias' for the prediction of NMR via understanding the query meaning.
Here, we introduce two technical contributions on how the SQuiD decides good or bad: (1) Co-occurrence table and (2) Learnable confounder.
%
%
Our experimental results show state-of-the-art performances and enhanced interpretability.

\section{Related Work}

\subsection{Video Moment Retrieval}

The video moment retrieval (VMR) is the task of finding a moment pertinent to a given natural language query.
The first attempts \cite{TALL,MCN} have been made to localize the moment 
by giving multi-modal feature interaction between query and video.
Previous VMR has focused on constructing modules that can help understand the contextual meaning of the query, including re-captioning \cite{EFRC} and temporal convolution \cite{yuan2019semantic}.
With the success of natural language models \cite{vaswani2017attention,liu2019roberta}, recent VMR systems are also interested in utilizing attention-based multi-modal interaction for the vision-language task.
%
Zhang et al. \cite{zhang2020span} conjugate question answering attention model into VMR as multi-modal span-based QA by treating the video as a text passage and target moment as the answer span. 
Wang et al. \cite{wang2021structured} perform multi-levels of cross-modal attention coupled with content-boundary moment interaction for accurate localization of moment.
Henceforth, there have been other efforts to perform a general format of video moment retrieval \cite{escorcia2019temporal,lei2020tvr,li2020hero}, which finds pertinent moments from a video corpus composed of multiple videos.
%
For this general VMR, Zhang et al. \cite{zhang2020hierarchical} suggested a hierarchical multi-modal encoder, which learns video and moment-level alignment for video corpus moment retrieval.
%
Zhang et al. \cite{zhang2021video} utilized multi-level contrastive learning to refine the alignment of text with video corpus, which enhances representation learning while keeping the video and text encoding separate for efficiency.
%
%
%
%
%
To advance forward general format of retrieval, we present another vulnerability as ``biased retrieval" in VMR and propose novel framework of debiasing to counter the retrieval bias.

\subsection{Causal Reasoning in Vision-Language}
Merged with natural language processing \cite{devlin2018bert,radford2018improving}, many high-level vision-language tasks have been introduced, including video/image captioning \cite{yu2016video,Venugopalan_2015_ICCV}, video moment retrieval \cite{gao2017tall,anne2017localizing}, and video/image question answering \cite{lei2018tvqa,tapaswi2016movieqa}.
%
Causal reasoning has recently contributed to another growth of these high-level tasks via giving ability to reason causal effects between vision and language modalities.
Wang \textit{et al.} \cite{wang2020visual} first introduced observational bias in visual representation and proposed to screen out confounding effect from the bias.
Recent question answering systems \cite{qi2020two,niu2021counterfactual} have also utilized this causal reasoning to eliminate language bias in question and answer.
For the moment retrieval, there have been efforts to remove the spurious correlation for correct retrieval \cite{nan2021interventional,yang2021deconfounded}. 
In this respect, we also uncover the retrieval bias, but furthermore, perform sensible debiasing by conjugating the bias in either positive or negative way.
%
\begin{figure*}[t]
    \centering
    \includegraphics[width=0.96\textwidth]{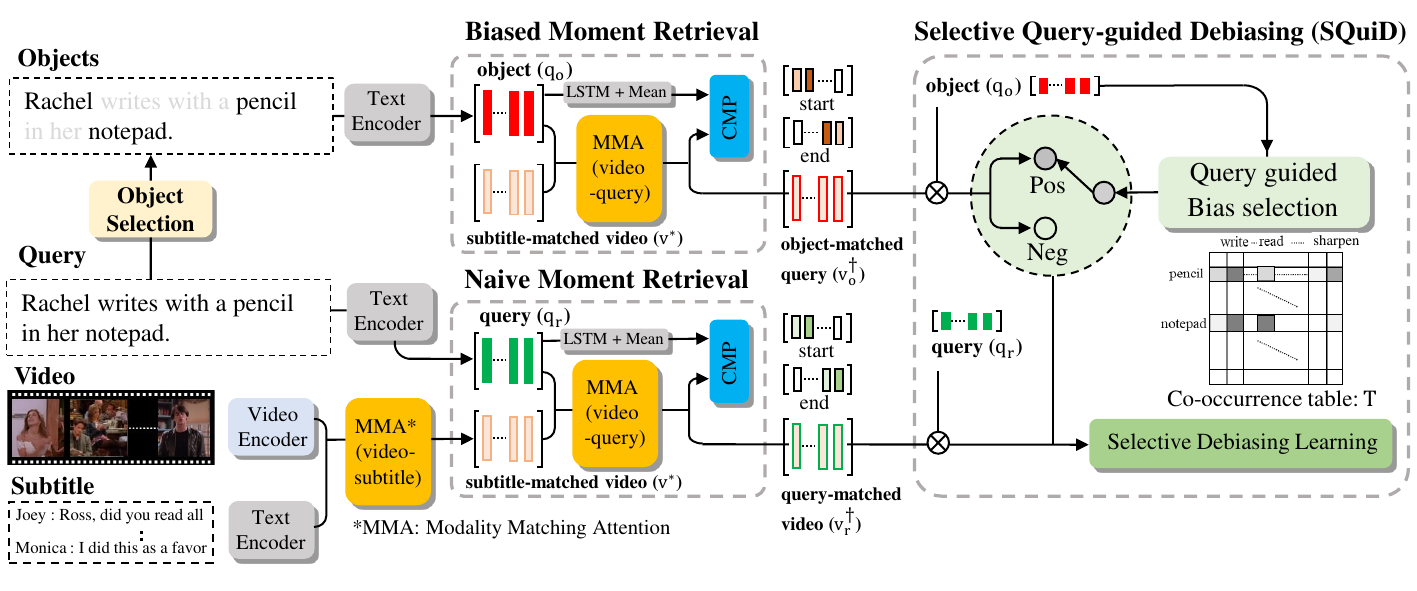}
    \caption{SQuiDNet is composed of 3 modules: (a) BMR which reveals biased retrieval, (b) NMR which performs accurate retrieval, (c) SQuiD which removes bad biases from accurate retrieval of NMR subject to the meaning of query.}
    \label{fig:3}
\end{figure*}

\section{Method}
\subsection{Selective Query-guided Debiasing Network}
Figure \ref{fig:3} illustrates Selective Query-guided Debiasing Network (SQuiDNet).
SQuiDNet prepares two moment retrievals under different motivations, where Naive Moment Retrieval (NMR) aims to perform accurate moment retrieval, while Biased Moment Retrieval (BMR) aims to explicitly reveal the retrieval bias in the training dataset.
Following, Selective Query-guided Debiasing (SQuiD) conjugates the biased prediction of BMR to selectively debias NMR.
Subject to the contextual meaning of the query, SQuiD decides positive or negative use of retrieval bias for contrastive learning between NMR and BMR.
To this, we present two technical contributions to the decision rule in SQuiD: (1) Co-occurrence table and (2) Learnable confounder.
%

\subsection{Input Representations}
SQuiDNet takes single pair of video (\ie video, subtitle) and query sentence as inputs, and training is performed under temporal boundary (\ie start-end time) annotations.
%
%
In inference, only video corpus and query are given, SQuiDNet predicts start-end time of moment pertinent to the query from the video corpus.
%
\paragraph{{\bf Video Representation.}}
We use 2D and 3D feature extractors for video encoder.
For 2D features, we use ResNet-101 \cite{he2016deep} pre-trained on ImageNet \cite{deng2009imagenet}, and for 3D features, we use SlowFast \cite{feichtenhofer2019slowfast} pre-trained on Kinetics \cite{kay2017kinetics}.
By concatenating the 2D and 3D features, $4352$-dimensional features $\mathbf{V} = \{\mathbf{v}_i\}_{i=1}^{N_\mathbf{v}}$ are used for video frame embedding, where $N_\mathbf{v}$ is number of frames in a video.
With $d$-dimensional embedder $\delta_{\mathbf{v}}$, final video features $v$ are embedded on top of layer normalization LN \cite{Ba_2016_arxiv} and positional encoding PE \cite{vaswani2017attention} as follows:
\begin{align}
    &v = \mbox{LN}(\delta_{\mathbf{v}}(\mathbf{V}) + \mbox{PE}(\mathbf{V})) \in \mathbb{R}^{N_{\mathbf{v}} \times d}.
\end{align}
\paragraph{{\bf Text Representation.}}
For text encoder, we use contextualized token embedding from pre-trained RoBERTa \cite{liu2019roberta}.
Here, we are given textual modalities as subtitle $\mathbf{S}=\{\mathbf{sub}(i)\}_{i}^{N_\mathbf{s}}$ and query $\mathbf{q}$, where $N_\mathbf{s}$ is the number of subtitles in a video.
We first tokenize all the words in subtitles and query into $5072$-dimensional word tokens, so that $\mathbf{W}_{\mathbf{sub}(i)} = \{\mathbf{w}_{\mathbf{sub}(i)}^{j}\}_{j=1}^{L_{{s}_{i}}}$ is word tokens in subtitle $\mathbf{sub}(i)$, where $L_{s_{i}}$ is number of words in that subtitle.
$\mathbf{W}_{\mathbf{q}} = \{\mathbf{w}_{\mathbf{q}}^{j}\}_{j=1}^{L_{\mathbf{q}}}$ is word token in query $\mathbf{q}$, where $L_{\mathbf{q}}$ is number of words in that query.
As like the video feature $v$, final subtitle $s_i$ and query $q_{r}$ features are embedded by $d$-dimensional embedder $\delta_{\mathbf{t}}$:
\begin{align}
    &s_i = \mbox{LN}(\delta_{\mathbf{t}}(\mathbf{W}_{\mathbf{sub}_{i}}) + \mbox{PE}(\mathbf{W}_{\mathbf{sub}_{i}}) \in \mathbb{R}^{L_{s_i} \times d},\\
    &q_{r} = \mbox{LN}(\delta_{\mathbf{t}}(\mathbf{W}_{\mathbf{q}}) + \mbox{PE}(\mathbf{W}_{\mathbf{q}})) \in \mathbb{R}^{L_{q} \times d},
    \label{eq:3}
\end{align}
\paragraph{{\bf Modality Matching Attention}}
As shown in Figure \ref{fig:3}, to give multi-modal interactions among input modalities (\ie video-subtitle, video-query), we define Modality Matching Attention (MMA) founded on multi-layer attention in Transformer \cite{Vaswani_2017_NIPS}.
MMA takes video and text as inputs and produces text-matched video features.
For mathematical definition of MMA, we first define $d$-dimensional input video feature $x = [x_{1}, \cdots, x_{n}] \in \mathbb{R}^{n \times d}$ and text features $y = [y_{1}, \cdots, y_{m}] \in \mathbb{R}^{m \times d}$, where $n,m$ is the number of video frames and words in the text. 
To give interactions between $x$ and $y$, we construct $z$ by concatenating $x$ and $y$ along the frame and word axis, and perform self-attention on $z$.
Here, we also add fixed token embedding $t_{<x>} \in \mathbb{R}^{n \times d}$ and $t_{<y>} \in \mathbb{R}^{m \times d}$ on $x$ and $y$, so that the Transformer identifies the heterogeneity between $x$ and $y$ as follows:
\begin{eqnarray}
    &x = x + t_{<x>}, y = y + t_{<y>},\\
    &z= [x||y] \in \mathbb{R}^{l \times d},\\
    &z^{\star}= \mbox{Self-Attention}(z) \in \mathbb{R}^{l \times d},\\
    &x^{\star}= \mbox{LN}(z^{\star}[:n] + x) \in \mathbb{R}^{n \times d},\\
    &\text{M}\text{MA}(x,y)= x^{\star},
\end{eqnarray}
where $[\cdot||\cdot]$ is concatenation and $l = n + m$ is the number of frames and words.
$[:]$ denotes slicing operation along the $l$ axis, such that we take video features $z^{\star}[:n]$ in $z^{\star}$ as text-matched video features $x^{\star}$.
Therefore, MMA produces $x^{\star} \in \mathbb{R}^{n \times d}$ comprehending language semantics in $y$.
Henceforth, we introduce MMA into two types of video-text matching: (1) video-subtitle matching and (2) video-query matching for following two retrieval models (\ie NMR and BMR).
\subsection{Biased Moment Retrieval and Naive Moment Retrieval}
Naive Moment Retrieval (NMR) is designed for the original purpose of moment retrieval, but Biased Moment Retrieval (BMR) aims at revealing retrieval bias.
Here, shown in Figure \ref{fig:3}, the beauty of our proposed BMR is its model-agnostic manner, following the identical structure of NMR.
In fact, NMR can be any model that performs moment retrieval (refer to experiments in Table \ref{tab:1}), and BMR serves to remove bias inherent in that NMR.
The only difference is that BMR takes object word features in query as inputs instead of full query sentence features.
As these object words lose the contextual meaning of original query, BMR can only depend on the object words to find the video moment, causing it to prioritize the moment that commonly appears together with that object.
To give mathematical definitions of NMR and BMR, we provide general formulations that can have variants of input text (\ie query or object words).
However, implementations of them should be independent, as they are trained for different purposes.
Below, the NMR and BMR are performed in the following process: (1) video-subtitle matching, (2) video-query matching, and (3) conditional moment prediction.

\paragraph{\bf Video-subtitle Matching}
Video frames and their subtitle appearing at the same time share common contextual semantics.
Motivated by video-subtitle matching in \cite{li2020hero}, we also introduce MMA on video frames and their shared subtitles to give multi-modal interactions among them.
For the inputs of MMA, we first reorganize video frames feature $v$ as video clips  $\mathbf{c} = \{c_i\}_{i=1}^{N_s}$ via collecting frames sharing single subtitle, where $c_i$ collects video frames that share $i$-th subtitle $s_{i}$. $N_s$ is the number of clips corresponding to the number of subtitles.
\begin{eqnarray}
    c_i^{\star} = \text{MMA}(c_i,s_i),
\end{eqnarray}
thus $c^{\star}_{i}$ represents $i$-th subtitle-matched video clip.
For the following video-query matching, we perform reunion of all clips $v^{\star} = c^{\star}_1 \cup \cdots \cup c^{\star}_{N_s}$ to reconstruct original frames and define $v^{\star} \in \mathbb{R}^{N_{v} \times d}$ as subtitle-matched video feature.
\paragraph{\bf Video-query Matching} As shown in Figure \ref{fig:3}, the $v^{\star}$ is utilized in MMA of two models (\ie NMR, BMR) for video-query matching.
But, for the input query, BMR utilizes object words instead of query sentence in order to learn retrieval bias.
To this, we use nouns from the query for object words as they manly contain objects.
Thus, we identify the part of speech (POS) of all words in query and sample noun words using natural language toolkit \cite{loper2002nltk} like below:
\begin{eqnarray}
    &\mathbf{W}_{\mathbf{o}} = \textrm{Noun}(\mathbf{W}_{\mathbf{q}}), \label{eq:noun}\\
    &q_{o} = \mbox{LN}(\delta_{\mathbf{t}}(\mathbf{W}_{\mathbf{o}}) + \mbox{PE}(\mathbf{W}_{\mathbf{o}})) \in \mathbb{R}^{L_{q_o} \times d},
    \label{eq:noun2}
\end{eqnarray}
where Noun($\cdot$) denotes noun-filtering operation using POS tagger.
The object words features $q_{o} \in \mathbb{R}^{L_{q_{o}} \times d}$ are also embedded from $\mathbf{W}_{o}$ like equation (\ref{eq:3}). 
The ${L_{q_{o}}}$ is number of objects in query.
Finally, we prepare query feature $q_{r}$ and object feature $q_{o}$ for video-query matching in NMR and BMR. 
Here, we define $q_{x} \in \{q_{r},q_{o}\}$ for general formulation of two models in following MMA.
The video-query matching is performed with $q_{x}$ and subtitle-matched video $v^{\star}$:
\begin{eqnarray}
    v^{\star \star} =  \text{MMA}(v^{\star},q_{x}),
    \label{eq:11}
\end{eqnarray}
where $v^{\star \star}$ is query-matched video, redefined as $v^{\dagger}_{x} = v^{\star \star}$ for two cases in $q_{x}$.
Thus, $v^{\dagger}_{x} \in \mathbb{R}^{N_v \times d}$ is our final video features for moment prediction with the query $q_{x}$.
\paragraph{\bf Conditional Moment Prediction} We predict the start-time and end-time of moment for moment prediction, where we introduce conditional moment prediction (CMP) under our motivation that one prediction (\eg start) can give causal information to the other prediction (\eg end) rather then predicting these two independently.
%
%
In details, given query feature $q_{x}$ and video feature $v^{\dagger}_{x}$, CMP first, predicts start-time of moment $t_{st}$.  
In here, we use query sentence feature $\mathbf{q}_{x} = \textrm{MeanPool}(\textrm{LSTM}(q_{x})) \in \mathbb{R}^{d \times 1}$ with lstm and mean-pooling over word axis to compute video-query similarities $v^{\dagger}_{x} \mathbf{q}_{x} \in \mathbb{R}^{N_v \times 1}$ in:
\begin{eqnarray}
    &P(t_{st}|v^{\dagger}_{x},q_{x}) = \mbox{Softmax}(\mbox{Conv1D}_{st}(v^{\dagger}_{x}\mathbf{q}_{x})) \in \mathbb{R}^{N_v \times 1},
    \label{eq:10}
\end{eqnarray}
where $\textrm{Conv1D}_{st}$ is 1D convolution layer to embed start-time information. 
After that, we predict end-time $t_{ed}$ with this prior start-time information $I_{st}$ below:
\begin{eqnarray}
    &I_{st} = \sigma(\mbox{Conv1D}_{st}(v^{\dagger}_{x}\mathbf{q}_{x})) \in \mathbb{R}^{N_v \times 1},\\
    &P(t_{ed}|v^{\dagger}_{x},q_{x}) = \mbox{Softmax}(\mbox{Conv1D}_{ed}(v^{\dagger}_{x}\mathbf{q}_{x} + \alpha I_{st})) \in \mathbb{R}^{N_v \times 1},
    \label{eq:14}
\end{eqnarray}
where $\sigma(\cdot)$ is nonlinear function like ReLU and $\alpha \in \mathbb{R}^{1}$ is learnable scalar.
These two predictions $P(t_{st}|v^{\dagger}_{x},q_{x})$ and $P(t_{ed}|v^{\dagger}_{x},q_{x})$ are trained from ground-truth start-end labels (\ie $g_{st},g_{ed}$) using cross-entropy loss $CE(\cdot,\cdot)$ as follows:
\begin{eqnarray}
    \mathcal{L}_{x} = CE(g_{st},P(t_{st}|v^{\dagger}_{x},\mathbf{q}_{x})) + CE(g_{ed},P(t_{ed}|v^{\dagger}_{x},\mathbf{q}_{x})).
    \label{loss}
\end{eqnarray}
Depending on subscript $x \in \{r,o\}$, BMR performs biased training from $\mathcal{L}_{o}$ and NMR performs retrieval training from $\mathcal{L}_{r}$.
Following SQuiD promotes selective debiasing NMR by conjugating retrieval bias in BMR.
%
%
\subsection{Selective Query-guided Debiasing}
Selective Query-guided Debiasing (SQuiD) is proposed to debias moment retrieval of NMR using biased retrieval from BMR.
SQuiD introduces contrastive learning to promote unbiased learning of NMR and biased learning of BMR, by contrasting the prediction of NMR as positive and BMR as negative.
However, biased predictions of BMR often should be positive for NMR, depending on the meaning of the query.
For example, when given a query as ``person drinks a coffee.", BMR also finds the scene of ``drinking coffee" in spite of input object words ``person" and ``coffee" due to spurious correlation between ``coffee " and ``drinking".
SQuiD needs to be sensible in determining whether to use retrieval bias as negative or positive according to the given query.
Therefore, our technical contribution is to introduce 2 different decision rules for SQuiD: (1) Co-occurrence table and (2) Learnable confounder.
\paragraph{{\bf Co-occurrence Table.}} Since we cannot directly know all the spurious correlations causing retrieval bias between `objects' and `scenes', we approximate these by referring to statistics of all query sentences.
We assume that the `predicate' in query would describe the `scene' in the video, so based on top-K (\eg K=100) frequent objects and predicates in training queries, we count the co-occurrence of predicates in query sentence for every object words.
This counting builds Co-occurrence table $T_{d} \in \mathbb{R}^{\textrm{K} \times \textrm{K}}$ of co-occurrence between object and predicate (Co-occurrence table is illustrated in SQuiD of Figure \ref{fig:3}).
The row in table $T_{d}$ holds the co-occurrence frequency of the predicates for a specific object.
Figure \ref{fig:2}(b) shows one of the co-occurrence distributions when the object ``television" is given.
To determine the biased prediction of BMR as negative or positive for contrastive learning with NMR, SQuiD utilizes the prior knowledge on predominant ``object-predicate" pairs in the Co-occurrence table.
For input object words in BMR, SQuiD identifies top-n (\eg n=10) predominant predicates in the Co-occurrence table.
If the top-n predicates appear in the original query sentence, SQuiD determines the prediction of BMR as positive instead of negative.
For the selective debiasing learning, we used hinged loss based on video-query similarity $v^{\dagger}_{x}\mathbf{q}_{x} \in \mathbb{R}^{N_{v}}$ between NMR and BMR as follows:
\begin{align}
\mathcal{L}_{hinge}^{n} = \mbox{max}[0,\Delta_{\mathbf{n}} - \mbox{max}(v^{\dagger}_{r}\mathbf{q}_{r}) + \mbox{max}(v^{\dagger}_{o}\mathbf{q}_{o})],\\
\mathcal{L}_{hinge}^{p} = \mbox{max}[0,\Delta_{\mathbf{p}} - \mbox{max}(v^{\dagger}_{r}\mathbf{q}_{r}) - \mbox{max}(v^{\dagger}_{o}\mathbf{q}_{o})]],
\end{align}
where $\mathcal{L}_{hinge}^{n}$ denotes the retrieval of BMR as negative and $\mathcal{L}_{hinge}^{p}$ denotes that as positive.
SQuiD's decision is to select one of them.
$\Delta_{\mathbf{n}}$ = 0.2 and $\Delta_{\mathbf{p}}$ = 0.4 is used and here, we give more margin for $\Delta_{\mathbf{p}}$ to promote learning of positive.
\paragraph{{\bf Learnable Confounder.}}
The Co-occurrence table can be a discrete approximation of retrieval bias as it assumes predefined predicates for selective debiasing.
For better approximation, we introduce learnable confounder $\mathbf{Z} \in \mathbb{R}^{\mathbf{K} \times d}$ that can learn object-scene spurious correlation, where it consists of $\mathbf{K}$ (\eg $\mathbf{K} = 100$) confounders with $d$-dimensional learnable parameters.
Assuming that predicate words sufficiently contain the contextual meaning of video scenes, we predict spuriously correlated \textit{predicate} feature $\mathbf{Y}_{B}$ from object words $q_{o}$ and the confounder $\mathbf{Z}$.
If the $\mathbf{Y}_{B}$ is similar to the predicate feature $\mathbf{Y}_{C}$ of the original query used in NMR, it means that predicate $\mathbf{Y}_{B}$ obtained from objects word and $\mathbf{Y}_{C}$ in given query have similar contextual meaning, thus, in this case, the retrieval of BMR should be used as positive. 
%

For above, we needs to pretrain $\mathbf{Z}$ to learn spurious correlations between objects and predicates, so that generated \textit{predicate} $\mathbf{Y}_{B}$ is biased predicate of the object.
To train $\mathbf{Z}$, we regress $\mathbf{Y}_{B}$ as $\mathbf{Y}_{C}$, which means $\mathbf{Z}$ is trained to generate predicate features $\mathbf{Y}_{B}$ that commonly appears together with given object words in a query.
To this, we first prepare mean-pooled objects feature $\mathbf{q}_{o} = \textrm{MeanPool}(\textrm{LSTM}(q_{o})) \in \mathbb{R}^{1 \times d}$ over word axis.
The $\mathbf{q}_{o}$ and confounder $\mathbf{Z}$ performs dot-product attention to make $\mathbf{Y}_{B}$, which regresses predicates feature $\mathbf{Y}_{C}$ in original query.
To get $\mathbf{Y}_{C}$, we sample predicate words $\mathbf{W}_{\mathbf{p}} = \textrm{Pred}(\mathbf{W}_{\mathbf{q}})$ in original query and embed predicate words feature $q_{p}$, which is the same process in equation (\ref{eq:noun},\ref{eq:noun2}) and Pred($\cdot$) denotes predicate-filtering.
%
%
%
\begin{align}
    \mathbf{Y}_{B} &= \mbox{Softmax}((\mathbf{q}_{\mathbf{o}}W_{\mathbf{o}})(\mathbf{Z}W_{\mathbf{z}})^{T})\mathbf{Z} \in \mathbb{R}^{d},\\
    \mathbf{Y}_{C}^{\star} &= \textrm{MeanPool}(\textrm{LSTM}(q_{p})) \in \mathbb{R}^{d},\\
    \mathcal{L}_{\mathbf{z}} &= ||\mathbf{Y}_{C}^{\star} - \mathbf{Y}_{B}||_{2}^{2}
    \label{eq:21}
\end{align}
where $W_{\mathbf{o}}, W_{\mathbf{z}} \in \mathbb{R}^{d \times d}$ are embedding matrices and $\mathbf{Y}_{C}^{\star}$ is fixed mean-pooled predicate features, which is target for L2 loss regression $\mathcal{L}_{\mathbf{z}}$.
%
%
%
After pretraining confounders $\mathbf{Z}$, SQuiD computes cosine similarity $r = \mbox{cosine}(\mathbf{Y}_{B},\mathbf{Y}_{C})$.
%
%
If $r$ is lager than 0, the retrieval from BMR is used as postive, otherwire as negative:
\begin{align}
    \mathcal{L}^{D}=\begin{cases}
\mathcal{L}^{n}_{hinge}&\text{if } r \leq 0\\
\mathcal{L}^{p}_{hinge}&\text{if } r > 0.
\end{cases}
\end{align}

\section{Experimental Results}
\begin{table*}[t]
	\centering
	\caption{Performances for video corpus moment retrieval on TVR (test-public), ActivityNet and DiDeMo. $\star$: reconstruction-based results, N: NMR, B: BMR}\smallskip
\begin{tabular}{l||ccc|ccc|ccc}
\Xhline{2\arrayrulewidth}
\hline
\multirow{3}{*}{Method} & \multicolumn{3}{c|}{TVR} & \multicolumn{3}{c|}{ActivityNet} & \multicolumn{3}{c}{DiDeMo (+ASR)}\\ \Xcline{2-10}{2\arrayrulewidth} 
& \multicolumn{3}{c|}{tIoU=0.7} & \multicolumn{3}{c|}{tIoU=0.7}& \multicolumn{3}{c}{tIoU=0.7} \\
                        & R@1   & R@10  & R@100 & R@1  & R@10 &R@100  & R@1  & R@10 &R@100  \\ \hline
                        \Xhline{2\arrayrulewidth}
XML \cite{lei2020tvr}  &3.32 &13.41 &30.52  &- &- &-   &$1.74^{\star}$ &$8.31^{\star}$ &$27.63^{\star}$  \\
HERO \cite{li2020hero} &6.21 &19.34 &36.66  &$1.19^{\star}$ &$6.33^{\star}$ &$16.41^{\star}$  &$1.59^{\star}$ &$9.12^{\star}$ &$29.23^{\star}$ \\
HAMMER \cite{zhang2020hierarchical} &5.13  &11.38 &16.71 &1.74 &8.75 &19.08  & - & - &- \\
ReLoCLNet \cite{zhang2021video} &4.15  &14.06 &32.42 &1.82 &6.91 &18.33  &- & - &-  \\
\hline
\textbf{SQuiDNet (N)}  &4.09 &12.30 &28.31 &1.62 &7.82 &18.53 &1.73 &9.84 &30.14 \\ 
\textbf{SQuiDNet (N\cite{li2020hero}, B)}  &8.34 &28.03 &35.45 &3.02 &10.23 &22.14 &2.62 &10.28 &31.11 \\
\textbf{SQuiDNet (N, B)}  &\textbf{10.09} &\textbf{31.22} &\textbf{46.05} &\textbf{4.43} &\textbf{12.81} &\textbf{26.54} &\textbf{3.52} &\textbf{12.93} &\textbf{34.03} \\\hline
\Xhline{2\arrayrulewidth}
\end{tabular}
\label{tab:1}
\end{table*}
\begin{table*}[t]
	\centering
	\caption{Performances for single video moment retrieval (SVMR) on TVR (val) and ActivityNet and video retrieval (VR) on TVR (val).}\smallskip
\begin{tabular}{l||cc|cc|l||cc}
\Xhline{2\arrayrulewidth}
\hline
\multirow{4}{*}{Method} &\multicolumn{4}{c|}{TVR} &\multirow{4}{*}{Method} &\multicolumn{2}{c}{ActivityNet}\\ \Xcline{2-5}{2\arrayrulewidth}\Xcline{7-8}{2\arrayrulewidth}
& \multicolumn{2}{c|}{SVMR} & \multicolumn{2}{c|}{VR} & & \multicolumn{2}{c}{SVMR}\\ \Xcline{2-5}{2\arrayrulewidth}\Xcline{7-8}{2\arrayrulewidth}
& \multicolumn{2}{c|}{R@1,tIoU=$\mu$}& \multicolumn{2}{c|}{-} & & \multicolumn{2}{c}{R@1,tIoU=$\mu$} \\
                        &$\mu$=0.5   &$\mu$=0.7  & R@1  & R@10 & &$\mu$=0.5  &$\mu$=0.7   \\ \hline
                        \Xhline{2\arrayrulewidth}
XML \cite{lei2020tvr}             &31.11  &13.89   & 16.54  & 50.41 &VSLNet \cite{zhang2020span} &43.22 &26.16   \\
HERO \cite{li2020hero}        & -  &4.02   & 30.11  & 62.69 &IVG \cite{nan2021interventional} &43.84 &27.1 \\
ReLoCLNet \cite{zhang2021video}   &31.88   &15.04   & 22.13  & 57.25 &SMIN \cite{wang2021structured} &48.46 &30.34 \\
\hline
\textbf{SQuiDNet (N,B)}   & \textbf{41.31} & \textbf{24.74} & \textbf{31.61} & \textbf{65.32} &\textbf{SQuiDNet} &\textbf{49.53} &\textbf{31.25} \\ \hline
\Xhline{2\arrayrulewidth}
\end{tabular}
\label{tab:2}
\end{table*}
\subsection{Dataset}
We validate SQuiDNet on three moment retrieval benchmarks as follows:
\paragraph{\bf{TV show Retrieval.}} TV show Retrieval (TVR) \cite{lei2020tvr} is composed of 6 TV shows across 3 genres: sitcoms, medical and crime dramas, which includes 109K queries from 21.8K multi-character interactive videos with subtitles. Each video is about 60-90 seconds in length. The TVR is split into $80\%$ train, $10\%$ val, $5\%$ test-private, $5\%$ test-public. The test-public is prepared for official challenge.

\paragraph{{\bf ActivityNet.}} ActivityNet Captions \cite{krishna2017dense} includes 20k videos with 100k query descriptions. 
10k videos are given for training and 5k for validation (val\_1), where the average length of all videos is 117 seconds, and the average length of queries is 14.8 words. 
We train our SQuiDNet and evaluate on the val\_1 split.

\paragraph{{\bf DiDeMo.}} The Distinct Describable Moments (DiDeMo) \cite{anne2017localizing} contains 10k videos under diverse scenarios. 
To mitigate complexity, most videos are about 30-seconds and uniformly divided into 5-seconds segments, thus a single video contains 21 possible segments.
DiDeMo is split into 80\% train, 10\% val, and 10\% test.
\subsection{Experimental Details}
\paragraph{{\bf Evaluation Metric.}}
We perform three retrieval tasks: (1) video retrieval (VR), (2) single video moment retrieval (SVMR), and (3) video corpus moment retrieval (VCMR).
VR is video-level retrieval, evaluating the number of correct predictions of video, where VR measures video-query similarities of all videos to select the highest one.
SVMR is moment-level retrieval in given video, evaluating the degree of overlap between predicted moments and ground-truth.
%
VCMR is moment-level retrieval in video corpus, thus we evaluate the incidences where: (1) the predicted video matches the ground-truth video; and (2) the predicted moment has high overlap with the ground-truth moment.
SQuiDNet predicts top-n (n=10) videos first, and performs moment retrieval on them.
%
Average recall at K (R@K) over query is used as the evaluation metric, where temporal Intersection over Union (tIoU) measures the overlap between predicted moment and the ground-truth.

%
%
%

\subsection{Results on Benchmarks}
Table \ref{tab:1} summarizes the best performances reported in XML \cite{lei2020tvr}, HERO \cite{li2020hero}, HAMMER \cite{zhang2020hierarchical}, ReLoCLNet \cite{zhang2021video} on TVR, ActivityNet and DiDeMo\footnote{please refer to Related Work and the papers for their detailed descriptions}. 
SQuiDNet outperforms previous state-of-the-art performance.
We also validate naive model without BMR, which shows large performance gap between full model of SQuiDNet, explaining the effectiveness of selective debiasing learning.
As SQuiDNet is conducted on model-agnostic manner, we replace NMR with the HERO baseline from their public code, which also shows improvement from original HERO.
%
SQuiDNet assumes subtitle as inputs, so we can utilize audio speech recognition (ASR) for DiDeMo, which is available in \cite{li2020hero}.
We also validate results without subtitle on DiDeMo by applying video feature $v$ instead of subtitle-matched video feature $v^{\star}$ in equation (9). 
This gives slight performance drop -0.36\%/-0.74\%/-1.82\% from full model of SQuiDNet, which explains that grouping video frames based on subtitles benefits understanding of contextual scenes in video.
Table \ref{tab:2} summarizes the results of two sub-tasks of VCMR: (1) SVMR and (2) VR on TVR and AtivityNet.
SQuiDNet also shows large performance gain on the SVMR and VR, which explains that selective debiasing is effective at both moment-level and video-level.
%
%
Although SQuiDNet is assumed to use subtitle, it also shows gain without subtitles in SVMR on ActivityNet.
%

\begin{table}
	\begin{minipage}{0.4\linewidth}
		\caption{Ablation on SQuiDNet variants for VCMR on TVR (val)}
		\label{tab:3}
		\centering
	    \begin{tabular}{l|| c}
		\Xhline{3\arrayrulewidth}
		 \multirow{2}{*}{Model variants}                           & tIoU=0.7 \\
		  & R@1\\
		\Xhline{2\arrayrulewidth}
		Full SQuiDNet                          & 8.52 \\ \hline
		\ \ w/o BMR     & 4.62 \\
		\ \ w/o CMP     & 8.17 \\
		\Xhline{2\arrayrulewidth}
		\ \ w/ (All negative)       & 6.41 \\
		\ \ w/ (All positive)       & 6.72 \\
		\ \ w/ (Co-occurrence table)     & 7.91 \\
		\ \ w/ (Learnable confounder)     & 8.52 \\
		\Xhline{3\arrayrulewidth}
	    \end{tabular}
	\end{minipage}\hfill
	\begin{minipage}{0.5\linewidth}
		\centering
		\includegraphics[width=60mm]{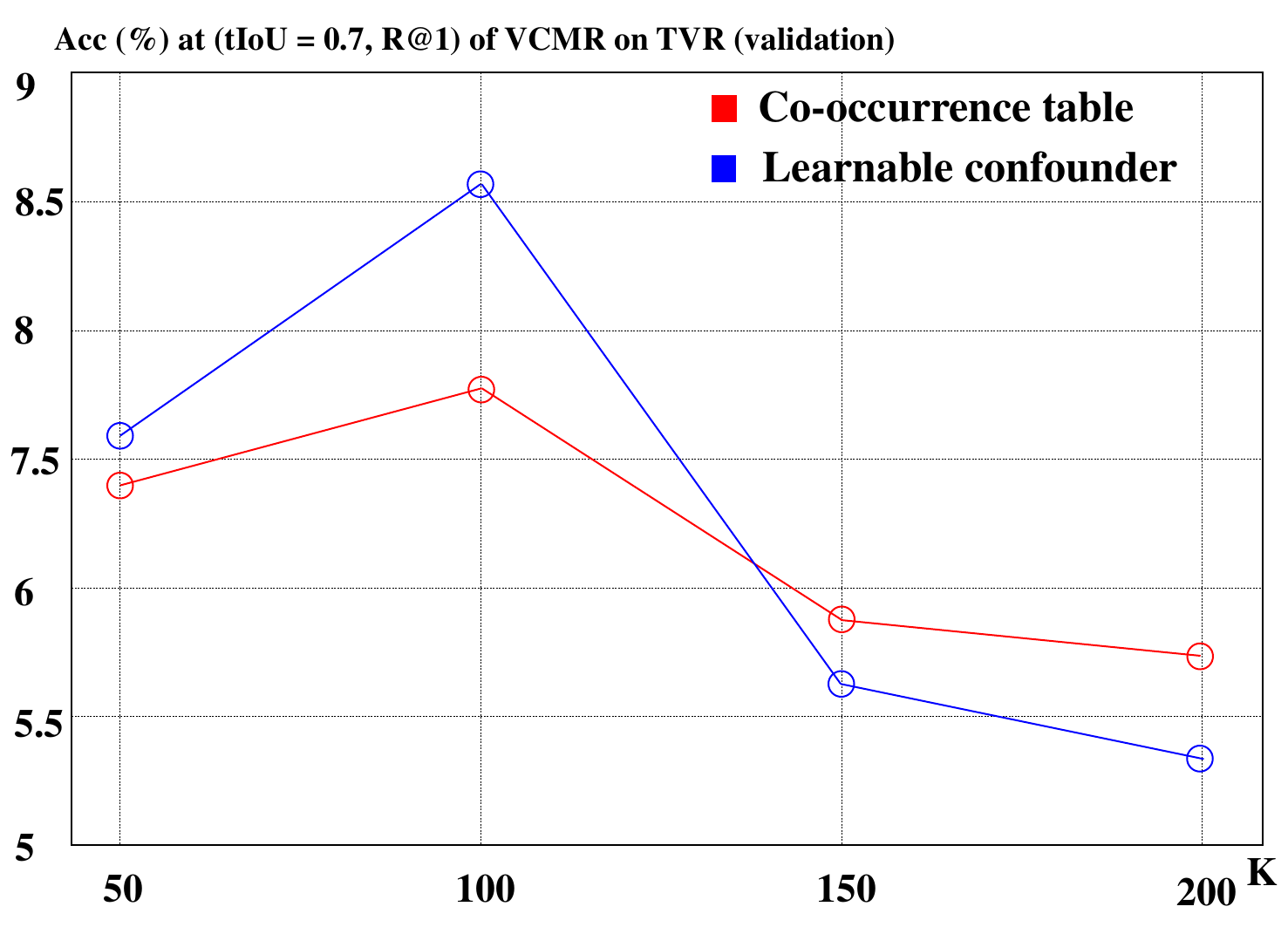}
		\captionof{figure}{Accuracy according to top-k objects for Co-occurrence table and variational k for Learnable confounder}
		\label{fig:4}
	\end{minipage}
\end{table}
%
%
\subsection{Ablation Study}

Table \ref{tab:3} presents ablation studies of our proposed components on SQuiDNet.
The first section reports full model SQuiDNet for VCMR on TVR validation set.
The second section shows ablative performance without Biased Moment Retrieval (BMR) and conditional moment prediction (CMP).
Large performance drop is shown without BMR, which gives two interpretations: (1) retrieval datasets contain many spurious correlations and (2) selective debiasing of video moment retrieval is non-trivial.
The performance gain of CMP is not as effective as that of BMR, however, it actually contributes on learning efficiency by 
promoting early convergence of training loss.
We consider this reason to be that CMP narrows the search space with prior knowledge of start-time.
%
%
%
Third section presents performance comparison from variants of SQuiD.
To be confident of variants of SQuiD's decision rule, we first conduct ablations of giving all retrievals from BMR as negative or positive for the contrastive learning with CMR. 
The result shows that positive use of biased retrieval is more effective than negative, which explains why SQuiD needs discernment on selecting biased retrieval.
Our proposed decision rules (\ie Co-occurrence table, Learnable confounder) give effectiveness via selective debiasing. 
The Learnable confounder is more effective, but it needs additional work to train the confouder $\mathbf{Z}$.
%
%

Figure \ref{fig:4} presents performances of VCMR according to the hyper-parameter $\mathbf{K}$ in Co-occurrence table and Learnable confounders.
For the Co-occurrence table, $\mathbf{K}$ is the number of top-k objects, and for Learnable confounder, $\mathbf{K}$ is the number of confounders.
Co-occurrence table utilizes statistics in training queries for approximating confounders while Learnable confounder learns confounder from object-predicate spurious correlations under our designed proxy learning with loss in equation (\ref{eq:21}), where both have similar curve.
However, Learnable confounder has higher best-performance.
We speculate Learnable confounder has superior control over confounders that cannot be defined in deterministic way.
\begin{figure}[t]
    \centering
    \includegraphics[width=\linewidth]{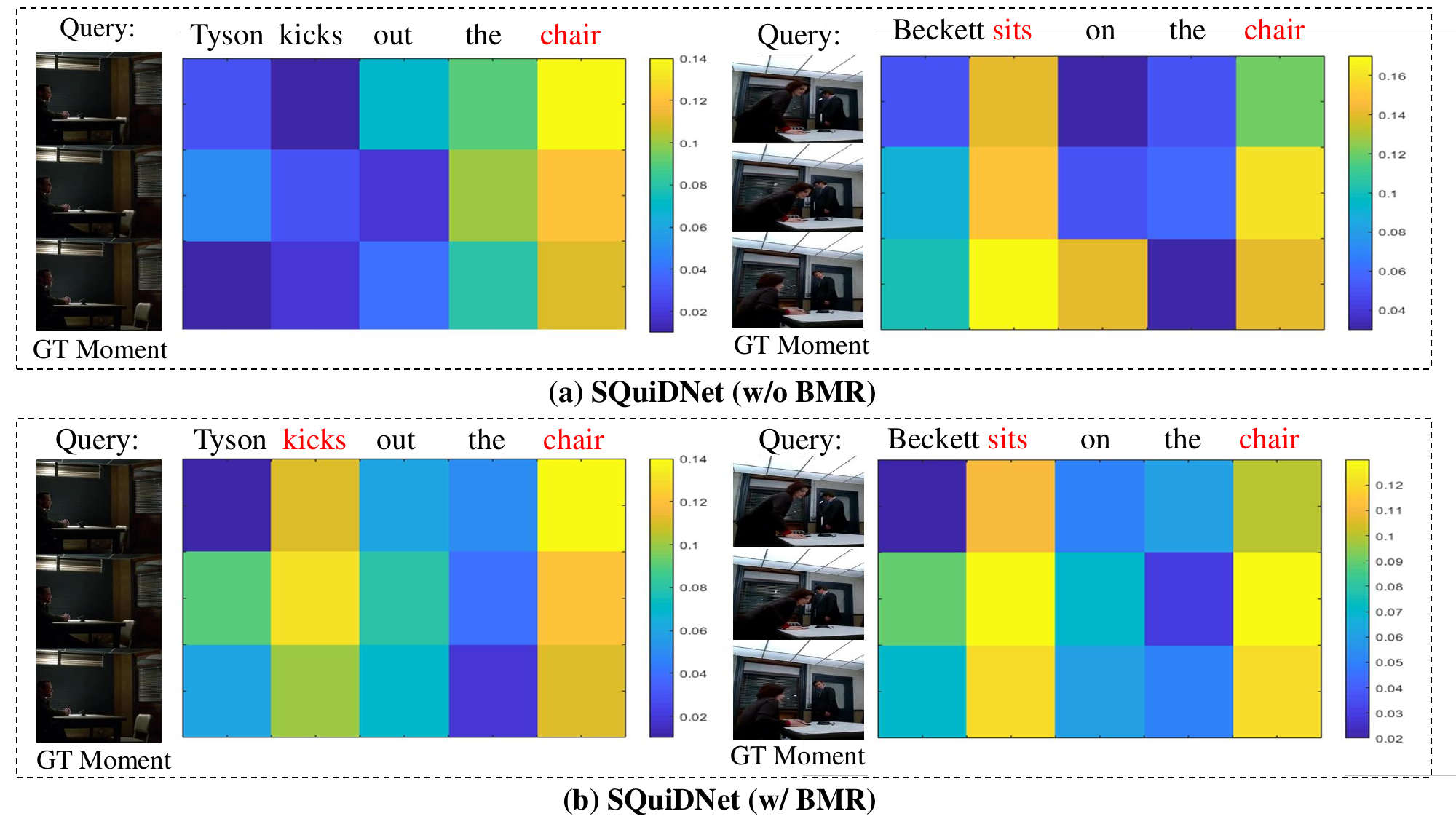}
    \caption{Visualization of word-level query-video similarities in GT moment. Upper box is results from SQuiDNet trained without BMR and lower box is results from SQuiDNet with BMR. It can be observed that the BMR enables the network to learn the uncommon predicate ``kicks" of the object ``chair" while also strengthens the learning of the spuriously correlated predicate ``sits."}
    \label{fig:5}
\end{figure}
\subsection{Qualitative Results}
Figure \ref{fig:5} shows the word-level query-video similarities when GT moment is given to SQuiDNet when two queries are given as ``Tyson kicks out the chair" and ``Beckett sits on the chair."
Figure \ref{fig:5}(a) represents the similarity distributions from SQuiDNet trained without BMR, where they show high similarity in word ``sit'' and low in ``kick." However in Figure \ref{fig:5}(b), the results with BMR show high similarities in both words ``sit'' and ``kick."
This explains that when one object word ``chair" is given, the system without debiasing can understand the spuriously correlated predicate word ``sit" but failed to learn the uncommon predicate word like ``kick."
In this respect, debiasing allows learning of object words' various connections with other predicate words. 
Furthermore, it can be observed that the system with debiasing also strengthens the understanding of the spuriously correlated predicate word by having higher similarities in more accurate moments. 

\begin{figure}[t]
    \centering
    \includegraphics[width=0.78\linewidth]{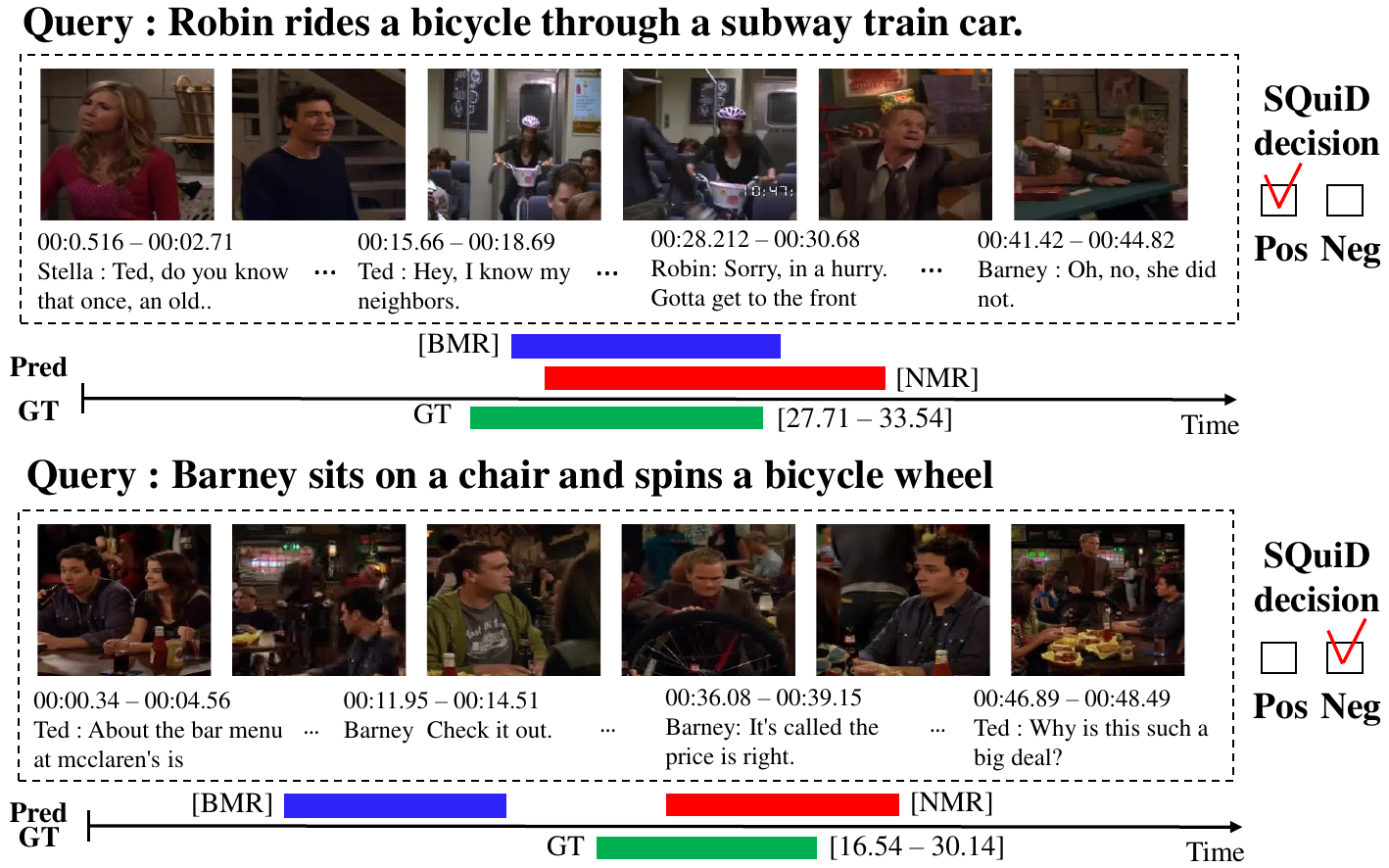}
    \caption{Visualization of moment prediction on NMR and BMR, where the SQuiD decision is represented for using retrieval bias from BMR as positive or negative.}
    \label{fig:6}
\end{figure}

Figure \ref{fig:6} presents moment predictions of the two models: NMR and BMR, where red box is prediction from NMR and blue box from BMR, while green box is ground-truth moment. 
In the right of the Figure, SQuiD's decision is shown on whether to use retrieval bias as positive or negative.
%
%
When the query is ``Robin rides a bicycle through a subway train car," both BMR and NMR predict the scene of person riding a bike in the video, where the SQuiD decides the prediction of BMR as positive retrieval bias.
But, for the query ``Barney sits on a chair and spins a bicycle wheel," BMR still predicts the scene of person riding a bike, which shows the retrieval bias between ``bicycle" and ``riding".
Here, to counter this retrieval bias, SQuiD decides the prediction of BMR as negative bias, such that NMR is trained to recede the bias prediction of BMR by contrastive learning. 
%
%

\section{Conclusion}

This paper considers Selective Query-guided Debiasing Network for video moment retrieval.
Although recent debiasing methods have focused on only removing retrieval bias, it sometimes should be preserved because there are many queries where biased predictions are rather helpful.
To conjugate this retrieval bias, SQuiDNet incorporates the following two main properties: (1) Biased Moment Retrieval that intentionally uncovers the biased moments inherent in objects of the query and (2) Selective Query-guided Debiasing that performs selective debiasing guided by the meaning of the query.
Our experimental results on three moment retrieval benchmarks (TVR, ActivityNet, DiDeMo) show effectiveness of SQuiDNet, while qualitative analysis shows improved interpretability.
\section*{Acknowledgemnet}
This work was partly supported by Institute for Information \& communications Technology Promotion(IITP) grant funded by the Korea government(MSIT) (No. 2021-0-01381, Development of Causal AI through Video Understanding) and partly supported by Institute of Information \& communications Technology Planning \& Evaluation(IITP) grant funded by the Korea government(MSIT) (No. 2022-0-00184, Development and Study of AI Technologies to Inexpensively Conform to Evolving Policy on Ethics)

\clearpage
%
%
\bibliographystyle{splncs04}
\bibliography{egbib}
\end{document}